\documentclass[11pt,a4paper]{article}
\usepackage[hyperref]{acl2018}
\usepackage{times}
\usepackage{latexsym}
\usepackage{graphicx,amsmath,bbm,array,float,amssymb}
\usepackage{subcaption}
\usepackage{enumitem}
\usepackage{url}

\aclfinalcopy 
\def\aclpaperid{1246} 
\usepackage{booktabs,chemformula}
\title{Large-Scale Multi-Domain Belief Tracking with Knowledge Sharing}
 \author{Osman Ramadan, Pawe{\l} Budzianowski, Milica Ga{\v{s}}i\'c \\
   Department of Engineering,\\
 University of Cambridge, U.K. \\
   {\tt \{oior2,pfb30,mg436\}@cam.ac.uk} }

\begin{document}
\maketitle
\begin{abstract}
 Robust dialogue belief tracking is a key component in maintaining good quality dialogue systems. The tasks that dialogue systems are trying to solve are becoming increasingly complex, requiring scalability to multi-domain, semantically rich dialogues. However, most current approaches have difficulty scaling up with domains because of the dependency of the model parameters on the dialogue ontology. In this paper, a novel approach is introduced that fully utilizes semantic similarity between dialogue utterances and the ontology terms, allowing the information to be shared across domains. The evaluation is performed on a recently collected multi-domain dialogues dataset, one order of magnitude larger than currently available corpora. Our model demonstrates great capability in handling multi-domain dialogues, simultaneously outperforming existing state-of-the-art models in single-domain dialogue tracking tasks.    
\end{abstract}

\section{Introduction}
Spoken Dialogue Systems (SDS) are computer programs that can hold a conversation with a human. These can be task-based systems that help the user achieve specific goals, e.g. finding and booking hotels or restaurants. In order for the SDS to infer the user goals/intentions during the conversation, its \emph{Belief Tracking} (BT) component maintains a distribution of states, called a \emph{belief state}, across dialogue turns~\cite{Steve:10}. 
The belief state is used by the system to take actions in each turn until the conversation is concluded and the user goal is achieved. In order to extract these belief states from the conversation, traditional approaches use a Spoken Language Understanding (SLU) unit that utilizes a semantic dictionary to hold all the key terms, rephrasings and alternative mentions of a belief state. The SLU then \emph{delexicalises} each turn using this semantic dictionary, before it passes it to the BT component~\cite{wang2013simple,henderson:14,williams2014web,zilka2015incremental,perez2016dialog, rast:17}. However, this approach is not scalable to multi-domain dialogues because of the effort required to define a semantic dictionary for each domain. More advanced approaches, such as the Neural Belief Tracker (NBT), use word embeddings to alleviate the need for \emph{delexicalisation} and combine the SLU and BT into one unit, mapping directly from turns to belief states ~\cite{Nikola:16}. Nevertheless, the NBT model does not tackle the problem of mixing different domains in a conversation. Moreover, as each slot is trained independently without sharing information between different slots, 
scaling such approaches to large multi-domain systems is greatly hindered. 

In this paper, we propose a model
that jointly identifies the domain and tracks the belief states corresponding to that domain. It uses semantic similarity between ontology terms and turn utterances to allow for parameter sharing between different slots across domains and within a single domain. In addition, the model parameters are independent of the ontology/belief states, thus the dimensionality of the parameters does not increase with the size of the ontology, making the model practically feasible to deploy in multi-domain environments without any modifications. Finally, we introduce a new, large-scale corpora of natural, human-human conversations providing new possibilities to train complex, neural-based models. 
Our model systematically improves upon state-of-the-art neural approaches both in single and multi-domain conversations.\\
\section{Background}\label{sec:back}

The belief states of the BT are defined based on an ontology - the structured representation of the database which contains entities the system can talk about. The ontology defines the terms over which the distribution is to be tracked in the dialogue. This ontology is constructed in terms of slots and values in a single domain setting. Or, alternatively, in terms of domains, slots and values in a multi-domain environment. Each domain consists of multiple slots and each slot contains several values, e.g. \texttt{domain=hotel}, \texttt{slot=price}, \texttt{value=expensive}. 
In each turn, the BT fits a distribution over the values of each slot in each domain, and a \emph{none} value is added to each slot to indicate if the slot is not mentioned so that the distribution sums up to 1. The BT then passes these states to the Policy Optimization unit as full probability distributions to take actions. This allows robustness to noisy environments~\cite{Steve:10}.
The larger the ontology, the more flexible and multi-purposed the system is, but the harder it is to train and maintain a good quality BT.


\section{Related Work}\label{sec:rel}
In recent years, a plethora of research has been generated on belief tracking ~\cite{williams:16}. For the purposes of this paper, two previously proposed models are particularly relevant.
\subsection{Neural Belief Tracker (NBT)}\label{subsec:nbt}
The main idea behind the NBT~\cite{Nikola:16} is to use semantically specialized pre-trained word embeddings to encode the user utterance, the system act and the candidate slots and values taken from the ontology. These are fed to semantic decoding and context modeling modules that apply a \emph{three-way} gating mechanism and pass the output to a non-linear classifier layer to produce a distribution over the values for each slot. It uses a simple update rule, $p(s_t) = \beta p(s_{t-1}) + \lambda y$, where $p(s_t)$ is the belief state at time step $t$, $y$ is the output of the binary decision maker of the NBT and $\beta$ and $\lambda$ are tunable parameters.
 
The NBT leverages semantic information from the word embeddings to resolve lexical/morphological ambiguity and maximize the shared parameters across the values of each slot. However, it only applies to a single domain and does not share parameters across slots. 

\subsection{Multi-domain Dialogue State Tracking}
Recently, ~\citet{rast:17} proposed a multi-domain approach using \emph{delexicalized} utterances fed to a two layer stacked bi-directional GRU network to extract features from the user and the system utterances. These, combined with the candidate slots and values, are passed to a feed-forward neural network with a softmax in the last layer. The candidate set fed to the network consists of the selected candidates from the previous turn and candidates from the ontology to a limit $K$, which restricts the maximum size of the chosen set. Consequently, the model does not need an ad-hoc belief state update mechanism like in the NBT.

The parameters of the GRU network are defined for the domain, whereas the parameters of the feed-forward network are defined per slot, allowing transfer learning across different domains. However, the model relies on \emph{delexicalization} to extract the features, which limits the performance of the BT, as it does not scale to the rich variety of the language. Moreover, the number of parameters increases with the number of slots.
\begin{figure*}
\begin{center}

	\includegraphics[width=0.9\linewidth]{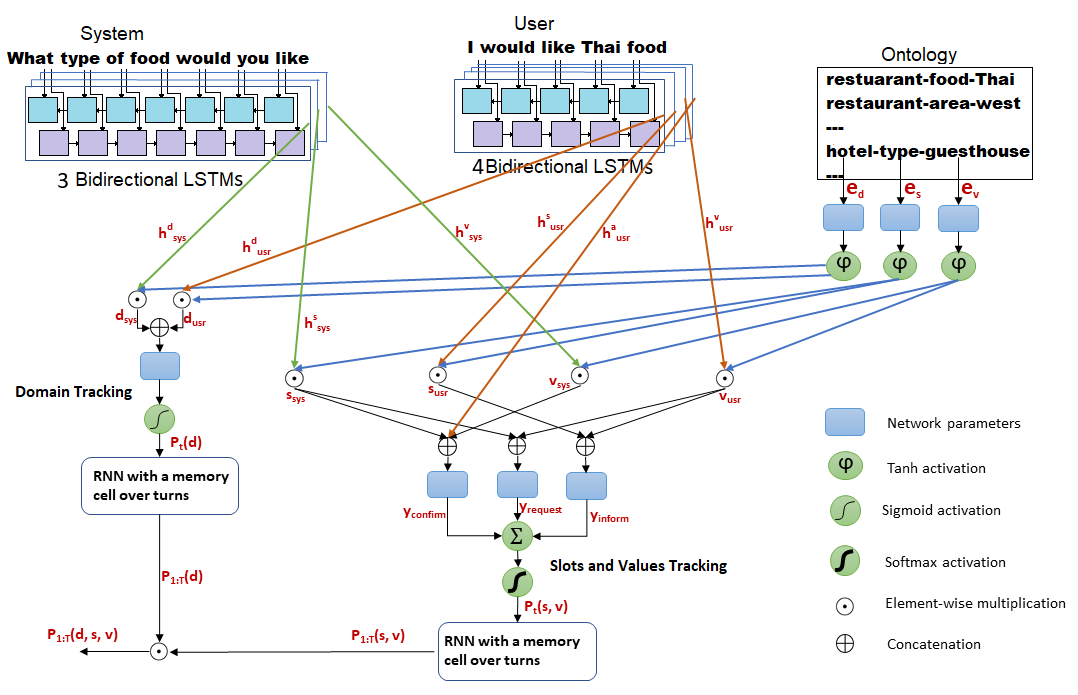}
\caption{The proposed model architecture, using Bi-LSTMs as encoders. Other variants of the model use CNNs as feature extractors~\cite{kim:14,kal:14}.}
\label{fig:arch}
\end{center}
\end{figure*}

\section{Method}\label{sec:method}
The core idea is to leverage semantic similarities between the utterances and ontology terms to compute the belief state distribution. In this way, the model parameters only learn to model the interactions between turn utterances and ontology terms in the semantic space, rather than the mapping from utterances to states. Consequently, information is shared between both slots and across domains. Additionally, the number of parameters does not increase with the ontology size. Domain tracking is considered as a separate task but is learned jointly with the belief state tracking of the slots and values. 
The proposed model uses semantically specialized pre-trained word embeddings~\cite{wiet:15}. 
To encode the user and system utterances, we employed $7$ independent bi-directional LSTMs \cite{graves2005framewise}. Three of them are used to encode the system utterance for domain, slot and value tracking respectively. Similarly, three Bi-LSTMs encode the user utterance while the last one is used to track the user affirmation. A variant of the CNNs as a feature extractor, similar to the one used in the NBT-CNN~\cite{Nikola:16} is also employed. 

\subsection{Domain Tracking}
Figure \ref{fig:arch} presents the system architecture with two bi-directional LSTM networks as information encoders running over the word embeddings of the user and system utterances.
The last hidden states of the forward and backward layers are concatenated to produce $\mathbf{h}_{usr}^{d}, \mathbf{h}_{sys}^{d}$ of size $L$ for the user and system utterances respectively. In the second variant of the model, CNNs are used to produce these vectors~\cite{kim:14,kal:14}. To detect the presence of the domain in the dialogue turn, element-wise multiplication is used as a similarity metric between the hidden states and the ontology embeddings of the domain: 
\begin{align*}
\mathbf{d}_k = \mathbf{h}_{k}^{d} \odot \text{tanh}(\mathbf{W}_{d} \ \mathbf{e}_d + \mathbf{b}_{d}),
\end{align*}
where $k \in \{usr, sys\}$, $\mathbf{e}_d$ is the embedding vector of the domain and $\mathbf{W}_{d} \in \mathcal{R}^{L\times D}$ transforms the domain word embeddings of dimension $D$ to the hidden representation. The information about semantic similarity is held by $\mathbf{d}_{usr}$ and $\mathbf{d}_{sys}$, which are fed to a non-linear layer to output a binary decision:
\begin{align*}
\mathcal{P}_{t}(d) = \sigma(\mathbf{w}_{d} \ \{\mathbf{d}_{usr} \oplus \mathbf{d}_{sys}\} + b_{d}),
\end{align*}
where $\mathbf{w}_{d} \in \mathcal{R}^{2L}$ and $b_{d}$ are learnable parameters that map the semantic similarity to a belief state probability $\mathcal{P}_{t}(d)$ of a domain $d$ at a turn {t}.

\subsection{Candidate Slots and Values Tracking}
Slots and values are tracked using a similar architecture as for domain tracking (Figure~\ref{fig:arch}).
However, to correctly model the context of the system-user dialogue at each turn, three different cases are considered when computing the similarity vectors:

\begin{enumerate}[nolistsep]
\item \textbf{Inform:}
The user is informing the system about his/her goal, e.g. \emph{'I am looking for a restaurant that serves Turkish food'}. 
\item \textbf{Request:}
The system is requesting information by asking the user about the value of a specific slot. If the system utterance is: \emph{'When do you want the taxi to arrive?'} and the user answers with \emph{'19:30'}.
\item \textbf{Confirm:}
The system wants to confirm information about the value of a specific slot. If the system asked: \emph{'Would you like free parking?'}, the user can either affirm positively or negatively. The model detects the user affirmation, using a separate bi-directional LSTM or CNN to output $\mathbf{h}_{usr}^{a}$.
\end{enumerate}
The three cases are modelled as following:
\begin{align*}
&y_{inf}^{s, v} = \mathbf{w}_{inf} \ \{\mathbf{s}_{usr} \oplus \mathbf{v}_{usr}\} + b_{inf},\\
&y_{req}^{s, v} = \mathbf{w}_{req} \ \{\mathbf{s}_{sys} \oplus \mathbf{v}_{usr}\} + b_{req},\\
&y_{af}^{s, v} = \mathbf{w}_{af} \ \{\mathbf{s}_{sys} \oplus \mathbf{v}_{sys} \oplus \mathbf{h}_{usr}^{a}\} + b_{af},
\end{align*}
where $\mathbf{s}_k, \mathbf{v}_k$ for $k \in \{usr, sys\} $ represent semantic similarity between the user and system utterances and the ontology slot and value terms respectively computed as shown in Figure~\ref{fig:arch}, and $\mathbf{w}$ and $b$ are learnable parameters.

\begin{table*}[h!]

\centering
\begin{tabular}{ c|c|c|c|c|c|c|}
\cline{2-7}
&\multicolumn{3}{c|}{WOZ 2.0}&\multicolumn{3}{c|}{New WOZ (only restaurants)}\\
\hline

\multicolumn{1}{|c|}{\textbf{Slot}} & \textbf{NBT-CNN} & \textbf{Bi-LSTM}& \textbf{CNN} & \textbf{NBT-CNN} & \textbf{Bi-LSTM} & \textbf{CNN}\\
 \hline
\multicolumn{1}{|c|}{Food}&88.9&96.1&\textbf{96.4}&78.3& 84.7 &\textbf{85.3}\\ 
\hline
\multicolumn{1}{|c|}{Price range}&93.7 &\textbf{98.0} &97.9 &92.6 &\textbf{95.6}  & 93.6\\ 
\hline
\multicolumn{1}{|c|}{Area}&94.3&97.8&\textbf{98.1}&78.3 & 82.6 & \textbf{86.4}\\ 
\hline
\multicolumn{1}{|c|}{Joint goals}&84.2&85.1&\textbf{85.5}&57.7 & 59.9 & \textbf{63.7}\\ 
\hline
\end{tabular}
\caption{WOZ 2.0 and new dataset test set accuracies of the NBT-CNN and the two variants of the proposed model, for slots \emph{food, price range, area} and \emph{joint goals.}}\label{tab:res1}

\hfill
\vspace*{-2em}
\end{table*}

The distribution over the values of slot $s$ in domain $d$ at turn $t$ can be computed by summing the unscaled states, $y_{inf}, y_{req}$ and $y_{af}$ for each value $v$ in $s$, and applying a softmax to normalize the distribution:
$$
\mathcal{P}_{t}(s, v) = \text{softmax}(y_{inf}^{s, v} + y_{req}^{s, v} + y_{af}^{s, v}).
$$
\subsection{Belief State Update}
Since dialogue systems in the real-world operate in noisy environments, a robust BT should utilize the flow of the conversation to reduce the  uncertainty in the belief state distribution. This can be achieved by passing the output of the decision maker, at each turn, as an input to an RNN that runs over the dialogue turns as shown in Figure~\ref{fig:arch}, which allows the gradients to be propagated across turns. This alleviates the problem of tuning hyper-parameters for rule-based updates. To avoid the vanishing gradient problem, three networks were tested: a simple RNN, an RNN with a memory cell \cite{henderson2014robust} and a LSTM. The RNN with a memory cell proved to give the best results. In addition to the fact that it reduces the vanishing gradient problem, this variant is less complex than an LSTM, which makes training easier. Furthermore, a variant of RNN used for domain tracking has all its weights of the form: $\mathbf{W}_i = \alpha_{i} \mathbf{I}$, where $\alpha_i$ is a distinct learnable parameter for hidden, memory and previous state layers and $\mathbf{I}$ is the identity matrix. Similarly, weights of the RNN used to track the slots and values is of the form:  $\mathbf{W}_j = \gamma_{j} \mathbf{I} + \lambda_{j}(\mathbf{1} - \mathbf{I})$, where $\gamma_{j}$ and $\lambda_{j}$ are the learnable parameters. These two variants of RNN are a combination of \citet{henderson2014robust} and \citet{Mrksic:18} previous works. The output is $\mathcal{P}_{1:T}(\mathbf{d})$ and $\mathcal{P}_{1:T}(\mathbf{s}, \mathbf{v})$, which represents the joint probability distribution of the domains and slots and values respectively over the complete dialogue. Combining these together produces the full belief state distribution of the dialogue:
$$
\mathcal{P}_{1:T}(\mathbf{d}, \mathbf{s}, \mathbf{v}) = \mathcal{P}_{1:T}(\mathbf{d}) \mathcal{P}_{1:T}(\mathbf{s}, \mathbf{v}).
$$
\subsection{Training Criteria}
Domain tracking and slots and values tracking are trained disjointly. Belief state labels for each turn are split into domains and slots and values. 
Thanks to the disjoint training, the learning of slot and value belief states are not restricted to a specific domain. Therefore, the model shares the knowledge of slots and values across different domains. The loss function for the domain tracking is:
$$
\mathcal{L}_d = -\sum_{n=1}^{N}\sum_{\mathbf{d} \in \mathcal{D}} t^{n}(\mathbf{d})\text{log} \mathcal{P}^{n}_{1:T}(\mathbf{d}),
$$
where $\mathbf{d}$ is a vector of domains over the dialogue, $t^{n}(\mathbf{d})$ is the domain label for the dialogue $n$ and $N$ is the number of dialogues. Similarly, the loss function for the slots and values tracking is: $$
\mathcal{L}_{s, v} = -\sum_{n=1}^{N}\sum_{\mathbf{s, v} \in \mathcal{S, V}} t^{n}(\mathbf{s, v})\text{log}\mathcal{P}^{n}_{1:T}(\mathbf{s, v}),
$$
where $\mathbf{s}$ and $\mathbf{v}$ are vectors of slots and values over the dialogue and $t^{n}(\mathbf{s, v})$ is the joint label vector for the dialogue $n$.

\section{Datasets and Baselines}\label{sec:data}
Neural approaches to statistical dialogue development, especially in a task-oriented paradigm, are greatly hindered by the lack of large scale datasets.
That is why, following the Wizard-of-Oz (WOZ) approach \cite{kelley1984iterative,wenN2N16}, we ran text-based multi-domain corpus data collection scheme through Amazon MTurk. The main goal of the data collection was to acquire human-human conversations between a tourist visiting a city and a clerk from an information center. At the beginning of each dialogue the user (visitor) was given explicit instructions about the goal to fulfill, which often spanned multiple domains. The task of the system (wizard) is to assist a visitor having an access to databases over domains.
The WOZ paradigm allowed us to obtain natural and semantically rich multi-topic dialogues spanning over multiple domains such as hotels, attractions, restaurants, booking trains or taxis. The dialogues cover from $1$ up to $5$ domains per dialogue greatly varying in length and complexity.
 
\subsection{Data Structure}
The data consists of $2480$ single-domain dialogues and $7375$ multi-domain dialogues usually spanning from $2$ up to $5$ domains. Some domains consists also of sub-domains like booking. The average sentence lengths are $11.63$ and $15.01$ for users and wizards respectively. The combined ontology consists of $5$ domains, $27$ slots and $663$ values making it significantly larger than observed in other datasets. To enforce reproducibility of results, we distribute the corpus with a pre-specified train/test/development random split. The test and development sets contain $1$k examples each. Each dialogues consists of a goal, user and system utterances and a belief state per turn. The data and model is publicly available.\footnote{http://dialogue.mi.eng.cam.ac.uk/index.php/corpus/}

\subsection{Evaluation}
We also used the extended WOZ 2.0 dataset \cite{wenN2N16}.\footnote{Publicly available at \url{https://mi.eng.cam.ac.uk/~nm480/woz_2.0.zip}.}  WOZ2 dataset consists of $1200$ single topic dialogues constrained to the restaurant domain. All the weights were initialised using normal distribution of zero mean and unit variance and biases were initialised to zero. ADAM optimizer ~\citep{kingma2014adam} (with 64 batch size) is used to train all the models for 600 epochs. Dropout~\cite{JMLR:v15:srivastava14a} was used for regularisation (50\% dropout rate on all the intermediate representations). 
For each of the two datasets we compare our proposed architecture (using either Bi-LSTM or CNN as encoders) to the NBT model\footnote{Publicly available at  \url{https://github.com/nmrksic/neural-belief-tracker}.} \cite{Nikola:16}.

\section{Results}\label{sec:result}
Table~\ref{tab:res1} shows the performance of our model in tracking the belief state of single-domain dialogues, compared to the NBT-CNN variant of the NBT discussed in Section~\ref{subsec:nbt}. Our model outperforms NBT in all the three slots and the joint goals for the two datasets. NBT  previously achieved state-of-the-art results~\cite{Nikola:16}. Moreover, the performance of all models is worse on the new dataset for restaurant compared to WOZ 2.0. This is because the dialogues in the new dataset are richer and more noisier, as a closer resemblance to real environment dialogues.   
\begin{table}
\centering
\begin{tabular}{ |c|c|c|}
\hline
\multicolumn{3}{|c|}{New WOZ (multi-domain)} \\
\hline
\textbf{Model} & \textbf{F1 score} & \textbf{Accuracy \%}  \\
 \hline
Uniform Sampling&0.108& 10.8 \\ 
\hline
Bi-LSTM&0.876  & \textbf{93.7}\\ 
\hline
CNN& \textbf{0.878} & 93.2 \\ 
\hline
\end{tabular}
\caption{The overall F1 score and accuracy for the multi-domain dialogues test set.\footnotemark\\
}\label{tab:res2}
\vspace{-2em}
\end{table}
\footnotetext{F1-score is computed by considering all the values in each slot of each domain as positive and the "none" state of the slot as negative.}

Table~\ref{tab:res2} presents the results on multi-domain dialogues from the new dataset described in Section~\ref{sec:data}. 
To demonstrate the difficulty of the multi-domain belief tracking problem, values of a theoretical baseline that samples the belief state uniformly at random are also presented. Our model gracefully handles such a difficult task. In most of the cases, CNNs demonstrate better performance than Bi-LSTMs. We hypothesize that this comes from the effectiveness of 
extracting local and position-invariant features, which are crucial for semantic similarities \cite{yin:17}.
\section{Conclusion}\label{sec:conc}
In this paper, we proposed a new approach that tackles the issue of multi-domain belief tracking, such as model parameter scalability with the ontology size. Our model shows improved performance in single-domain tasks compared to the state-of-the-art NBT method. By exploiting semantic similarities between dialogue utterances and ontology terms, the model alleviates the need for ontology-dependent parameters and maximizes the amount of information shared between slots and across domains. In future, we intend to investigate introducing new domains and ontology terms without further training thus performing zero-shot learning. 

\section*{Acknowledgments}
The authors would like to thank Nikola Mrk{\v{s}}i\'c, Jacquie Rowe, the Cambridge Dialogue Systems Group and the ACL reviewers for their constructive feedback. Pawe\l{} Budzianowski  is  supported  by  EPSRC  Council and Toshiba Research Europe Ltd, Cambridge Research  Laboratory. The data collection was funded through Google Faculty Award.

\bibliography{acl2018}
\bibliographystyle{acl_natbib}

\end{document}